\documentclass{adobe_research}
\usepackage[T1]{fontenc}
\usepackage{lmodern}
\usepackage{multirow}
\usepackage{soul}
\usepackage{pifont}
\usepackage[T1]{fontenc}

\usepackage{graphicx}
\usepackage{amsmath,amssymb}

\usepackage{makecell}
\usepackage[dvipsnames]{xcolor}
\usepackage{color, colortbl}
\usepackage{caption}
\usepackage{algorithm}

\newcommand{\Method}{Method}

\usepackage{xspace}
\makeatletter
\DeclareRobustCommand\onedot{\futurelet\@let@token\@onedot}
\def\@onedot{\ifx\@let@token.\else.\null\fi\xspace}

\def\eg{\emph{e.g}\onedot}

\makeatother

\let\cite\citep

\usepackage{newtxtext}

\usepackage{algpseudocode}
\definecolor{catgray}{gray}{0.92}

\usepackage[dvipsnames]{xcolor}
\definecolor{FutureOrange}{HTML}{EC866D}

\usepackage{booktabs}
\usepackage{wrapfig}
\usepackage{float}

\title{WorldCam: Interactive Autoregressive 3D Gaming Worlds with Camera Pose as a Unifying Geometric Representation}

\author[1*]{Jisu Nam}
\author[2]{Yicong Hong}
\author[2]{Chun-Hao Paul Huang}
\author[2]{Feng Liu}
\author[1]{JoungBin Lee}
\author[1]{Jiyoung Kim}
\author[1]{Siyoon Jin}
\author[3]{Yunsung Lee}
\author[3]{Jaeyoon Jung}
\author[3]{Suhwan Choi}
\author[1\dagger]{Seungryong Kim}
\author[2\dagger]{Yang Zhou}

\affiliation[1]{KAIST}
\affiliation[2]{Adobe Research}
\affiliation[3]{MAUM AI}

\contribution[\dagger]{Co-corresponding Authors}
\contribution[*]{Work done during Adobe Research internship.}

\abstract{
Recent advances in video diffusion transformers have enabled interactive gaming world models that allow users to explore generated environments over extended horizons. However, existing approaches struggle with precise action control and long-horizon 3D consistency. Most prior works treat user actions as abstract conditioning signals, overlooking the fundamental geometric coupling between actions and the 3D world, whereby actions induce relative camera motions that accumulate into a global camera pose within a 3D world. In this paper, we establish camera pose as a unifying geometric representation to jointly ground immediate action control and long-term 3D consistency. First, we define a physics-based continuous action space and represent user inputs in the Lie algebra to derive precise 6-DoF camera poses, which are injected into the generative model via a camera embedder to ensure accurate action alignment. Second, we use global camera poses as spatial indices to retrieve relevant past observations, enabling geometrically consistent revisiting of locations during long-horizon navigation. To support this research, we introduce a large-scale dataset comprising 3,000 minutes of authentic human gameplay annotated with camera trajectories and textual descriptions. Extensive experiments show that our approach substantially outperforms state-of-the-art interactive gaming world models in action controllability, long-horizon visual quality, and 3D spatial consistency. 
}

\date{2026.03.17}

\adobedata[Project Page]{\url{https://cvlab-kaist.github.io/WorldCam/}}

\begin{document}

\maketitle

\begin{figure}[!h]
    \centering
    \vspace{-6mm}
    \includegraphics[width=1\textwidth]{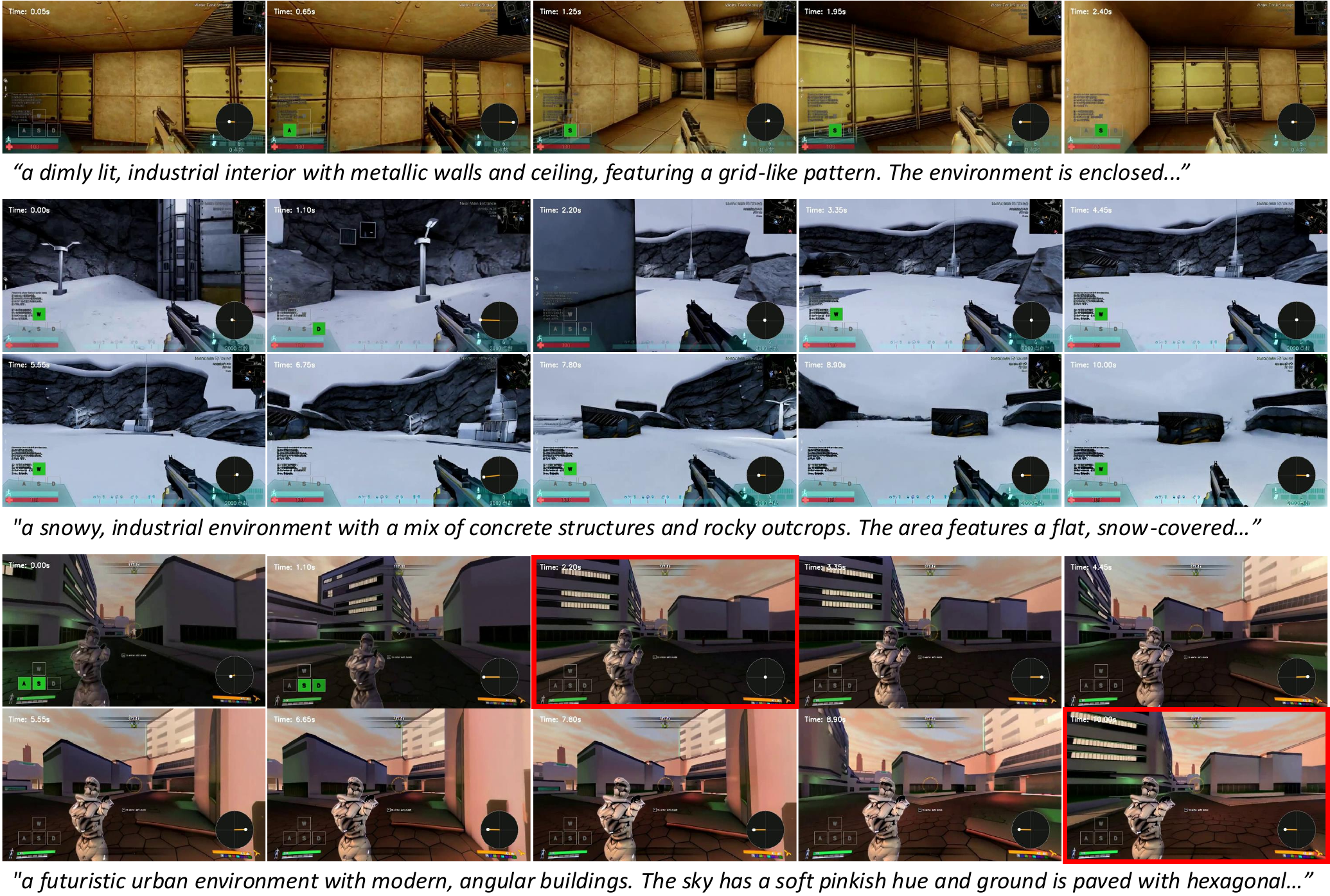}
    \vspace{-6mm}
    \caption{\textbf{Teaser (Best viewed in color and zoomed in):} WorldCam is an interactive 3D gaming model that enables \textbf{precise action control under challenging keyboard and mouse inputs (top), supports long-horizon interactions (middle), and preserves consistent 3D geometry across viewpoints (bottom).} Time (seconds at 20 FPS) is visualized in the top-left of each frame, while keyboard and mouse inputs are shown in the bottom-left and bottom-right, respectively. The red box highlights consistent 3D geometry in revisited views.}
    \label{fig:teaser}
    \vspace{-4mm}
\end{figure}

\section{Introduction}

Recent advances in Video Diffusion Transformers (DiTs)~\cite{yang2024cogvideox, wan2025wan, hacohen2024ltx, agarwal2025cosmos} have significantly improved the realism and scalability of video generation. Building on this progress, recent studies~\cite{valevski2024diffusion, che2024gamegen, bruce2024genie, zhang2025matrix, he2025matrix, gao2025adaworld, chen2025learning} take important steps toward \textit{interactive gaming world models}, demonstrating that generative models can simulate playable environments. However, despite plausible visual outputs, they still struggle with precise action control and 3D world consistency, which are prerequisites for a functional gaming engine.

This limitation arises because prior works have overlooked the \textbf{fundamental geometric coupling between user actions and the 3D world}. In gaming environments, user actions (e.g., keyboard presses and mouse movements) are not abstract control signals, but instead induce \textit{relative camera motions within a 3D scene}. These relative motions accumulate over time to form the camera’s global trajectory, which dictates how the underlying 3D world is projected into 2D observations. As a result, accurate action control and 3D consistency are not independent objectives, but are inherently coupled through the camera pose.

Despite this geometric coupling, existing interactive gaming world models~\cite{valevski2024diffusion, che2024gamegen, bruce2024genie, zhang2025matrix, he2025matrix, gao2025adaworld, chen2025learning} treat actions as abstract conditioning signals by directly injecting raw action inputs into video generative models. This design leads to misaligned camera motion and inconsistent 3D geometry due to the lack of explicit geometric constraints. Several camera-controlled video generation methods~\cite{wang2024motionctrl, he2024cameractrl, wan2025wan} address 3D consistency by conditioning generation on camera poses, however, they focus on short videos (e.g., 16 frames) and fail to model action-driven control and long-horizon inference.

In this paper, we introduce \textbf{WorldCam}, a foundation model for interactive gaming worlds built on a video DiT backbone. WorldCam enables precise action control, long-horizon inference, and consistent 3D world modeling. Our core contribution lies in establishing \textbf{camera pose as a unifying geometric representation} that jointly grounds both immediate action control and long-horizon 3D consistency.

Compared to prior works that directly inject raw action signals into video generative models, we define a physics-based continuous action space that translates complex user action inputs (e.g., coupled keyboard and mouse actions) into geometrically accurate camera poses. Unlike prior approaches~\cite{li2025hunyuan} relying on naive, decoupled linear approximations, which inherently fail to capture coupled dynamics such as screw motion, we model user actions as spatial velocities in the Lie algebra~\cite{hall2013lie} and strictly derive precise 6-DoF relative poses. These poses are then encoded as Plücker embeddings~\cite{sitzmann2021light} via a camera embedder and injected into intermediate features of the video DiT, ensuring that the generated outputs adhere to the intended physical motion.

Crucially, the camera pose serves a dual purpose, acting not only as a control signal for actions but also as an explicit geometric cue for long-horizon 3D consistency. Specifically, we retrieve relevant previously generated latents based on camera pose similarity between the current and past camera poses. These latents are concatenated with the current latent sequence, and their associated camera pose embeddings establish geometric correspondences between the current latents and retrieved past latents. By grounding both action control and 3D geometry in a shared camera space, we achieve precise action control and long-horizon 3D world consistency.

A major obstacle to building interactive gaming world models is the lack of large-scale, high-fidelity video datasets that capture real human gameplay dynamics. Existing works often rely on Minecraft datasets~\cite{chen2025learning, guo2025mineworld, po2025long, zhang2025matrix}, which are limited by simplified geometry and discrete motion patterns, or on closed-licensed game video datasets~\cite{li2025hunyuan, tang2025hunyuan, bruce2024genie} that are inaccessible for reproducible research.

To address this, we introduce \textbf{WorldCam-50h}, a large-scale dataset comprising 3,000 minutes of authentic human gameplay collected from one closed-licensed commercial game, \textit{Counter-Strike}, and two open-licensed games, \textit{Xonotic} and \textit{Unvanquished}. To capture diverse human player behaviors, the dataset covers complex scenarios such as general navigation, rapid 360° camera rotations, and reverse traversal across varied geometries. Furthermore, each video is annotated with rich textual descriptions~\cite{yang2025qwen3} and pseudo ground-truth camera poses~\cite{huang2025vipe}.

Through extensive experiments, we demonstrate that our approach simultaneously achieves precise action alignment, sustained long-horizon visual quality, and robust 3D spatial consistency. WorldCam consistently outperforms prior interactive gaming world models~\cite{he2025matrix, mao2025yume, li2025hunyuan} as well as camera-controllable video generation methods~\cite{wang2024motionctrl, wan2025wan}. We validate the effectiveness of our design choices through comprehensive ablation studies. We will publicly release our open-licensed datasets, code, and pretrained models.

\begin{figure*}[!t]
    \centering
    \includegraphics[width=1\textwidth]{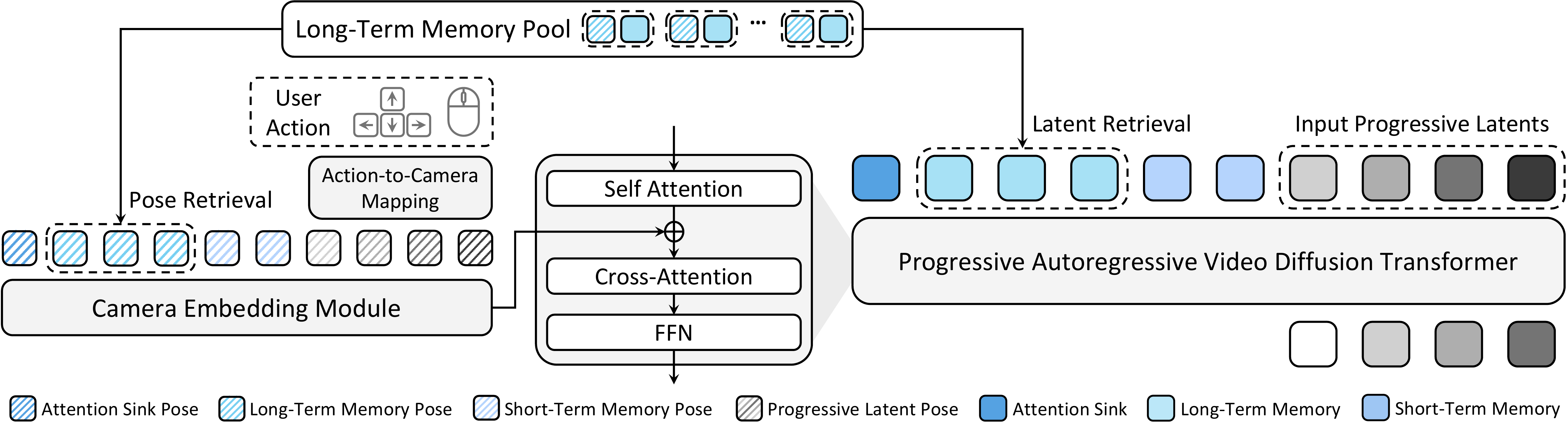}
    \vspace{-15pt}
   \caption{\textbf{Overall architecture.} WorldCam converts user actions into camera poses in Lie algebra and conditions a progressive autoregressive video transformer on these camera poses for precise action control. Retrieved long-term memory latents and camera poses from the memory pool enforce 3D consistency of the generated world, while short-term memory with an attention sink stabilizes long-horizon generation.}

    \label{fig:architecture}
\end{figure*}

\begin{table*}[!t]
\centering
\caption{\textbf{Comparison with existing interactive gaming world models and camera-controlled models.}}
\vspace{-5pt}
\label{tab:model}

\resizebox{\textwidth}{!}{
\begin{tabular}{l|cccccc|cc|c}
\hline
 & The Matrix & Matrix-game & Matrix-game 2.0 & Genie & Yan & GameCraft & CameraCtrl & MotionCtrl
 & \textbf{WorldCam}\\
\hline
Action control        & $\checkmark$  & $\checkmark$ & $\checkmark$ & $\checkmark$ & $\checkmark$ & $\checkmark$ & $\times$ & $\times$ & $\checkmark$ \\
3D consistency        & $\times$ & $\times$ & $\times$ & $\times$ & $\times$ & $\times$ & $\checkmark$  & $\checkmark$  & $\checkmark$ \\
Long-horizon inference& $\times$ & $\times$ & $\checkmark$ & $\checkmark$ & $\checkmark$ & $\checkmark$ & $\times$ & $\times$ & $\checkmark$ \\
\hline
\end{tabular}
}
\end{table*}

\section{Related Work}

\paragraph{\normalfont\bfseries Interactive gaming world models.}
Prior works typically inject raw action signals into video generative models via cross-attention~\cite{feng2024matrix, he2025matrix, valevski2024diffusion}, AdaLN~\cite{xiao2025worldmem}, or text~\cite{mao2025yume, chen2025deepverse}. However, by treating actions as abstract conditioning signals, they often fail to model accurate camera motion and achieve 3D consistency, since raw actions lack an understanding of the underlying 3D scene. GameCraft~\cite{li2025hunyuan} improves action control by linearly approximating user inputs into camera poses, but ignores the underlying $SE(3)$ geometry by decoupling translation and rotation. This makes it struggle to capture coupled dynamics such as screw motion arising from entangled keyboard and mouse inputs. Moreover, camera motion in GameCraft is used only for immediate control, lacking a persistent geometric anchor for 3D consistency. In contrast to prior works that fail to simultaneously satisfy precise action control, long-horizon inference, and 3D consistency, we propose a camera-grounded framework that uses camera pose as a unified geometric representation to achieve all three (Table~\ref{tab:model}).

\vspace{-5pt}
\paragraph{\normalfont\bfseries Interactive gaming datasets.}
Large-scale gaming video datasets that capture authentic human gameplay motion are crucial for training interactive gaming world models. Prior works~\cite{chen2025learning, guo2025mineworld, po2025long, zhang2025matrix} commonly rely on Minecraft dataset~\cite{guss2019minerl}, which provides paired action labels but suffers from limited visual diversity and simplified geometry. Recent efforts~\cite{tang2025hunyuan, li2025hunyuan} utilize internal, closed-licensed gameplay datasets, hindering research reproducibility. In contrast, we introduce a large-scale, open-licensed dataset capturing diverse and dynamic human gameplay, fully annotated with pseudo ground-truth camera poses and textual descriptions.

\section{WorldCam}
\label{sec:worldcam}

Given an initial RGB observation $I_0$ (a single image or a short video clip), a text prompt $c_{\text{text}}$, and a sequence of user actions $\{A_1, A_2, \ldots\}$, our goal is to autoregressively generate a sequence of video frames $\{I_1, I_2, \ldots\}$ that (i) accurately follow the user actions (Section~\ref{sec:action_control} \&~\ref{sec:camera_control}), (ii) remain consistent with a single 3D world (Section~\ref{sec:3d_consistency}), and (iii) maintain high visual quality over long horizons (Section~\ref{sec:autoregressive}).

Table~\ref{tab:model} compares our method with recent interactive world models~\cite{feng2024matrix, zhang2025matrix, he2025matrix, bruce2024genie, ye2025yan, li2025hunyuan} and camera-controlled approaches~\cite{wan2025wan, wang2024motionctrl}. Figure~\ref{fig:architecture} provides an overview of the overall architecture.

\subsection{Baseline: Video Diffusion Transformer}
\label{sec:baseline}
We build WorldCam on a pretrained video Diffusion Transformer (DiT), Wan-2.1-T2V~\cite{wan2025wan},
which consists of spatio-temporal self-attention and text cross-attention layers. Given an input video ${V} \in \mathbb{R}^{F \times H \times W \times 3}$, the VAE encoder maps it to a latent sequence
$\mathbf{z}_0 \in \mathbb{R}^{f \times h \times w \times c}$.
Here, $(F, H, W, 3)$ denote the number of frames, height, width, and RGB channels of the video, while $(f, h, w, c)$ denote the corresponding dimensions in the latent space.

Gaussian noise ${\epsilon} \sim \mathcal{N}({0}, {I})$ is added to ${z}_0$ to obtain noisy latents
$\mathbf{z}_t \in \mathbb{R}^{f \times h \times w \times c}$ at timestep $t \in [0, T]$. Given a text prompt $c_\text{text}$, the DiT ${v}_\theta$ learns to predict the velocity field that transports $\mathbf{z}_t$ toward $\mathbf{z}_0$~\cite{lipman2022flow}. The training objective is defined as
\begin{equation}
\boldsymbol{L}_{\text{FM}} =
\mathbb{E}_{\mathbf{z}_0, c_\text{text}, t}
\Bigl[
\bigl\|
{v}_\theta(\mathbf{z}_t, c_\text{text}, t) - \frac{\mathbf{z}_0 - \mathbf{z}_t}{1 - t}
\bigr\|_2^2
\Bigr].
\end{equation}
 
\subsection{Action-to-Camera Mapping}
\label{sec:action_control}
A central challenge in interactive 3D world modeling is to translate user actions into physically consistent camera motion. Prior works often directly inject raw action signals into the generative model~\cite{he2025matrix, feng2024matrix, che2024gamegen}, rely on text prompts to describe actions~\cite{mao2025yume}, or adopt linear action-to-camera approximations~\cite{li2025hunyuan}.  Such formulations frequently lead to misaligned or geometrically inconsistent camera trajectories, especially under complex motions involving coupled translation and rotation.

To ensure physically accurate and fine-grained action control, we define the action space in the Lie algebra $\mathfrak{se}(3)$. At each transition from frame $I_{i-1}$ to $I_i$, the user action $A_i$ is represented as a twist vector $A_i = [\mathbf{v}_i; \boldsymbol{\omega}_i] \in \mathbb{R}^6$, where $\mathbf{v}_i = [v_x, v_y, v_z]^\top \in \mathbb{R}^3$ and $\boldsymbol{\omega}_i = [\omega_x, \omega_y, \omega_z]^\top \in \mathbb{R}^3$ denote the linear and angular velocities, respectively. We then derive the corresponding relative camera pose $\Delta P_i \in SE(3)$ via the matrix exponential map: \begin{equation} \Delta P_i = \exp(\hat{A}_i) = \begin{bmatrix} \Delta R_i & \Delta t_i \\ \mathbf{0}^\top & 1 \end{bmatrix}, \label{eq:lie_mapping} \end{equation} where $\hat{A}_i \in \mathfrak{se}(3)$ denotes the $4 \times 4$ matrix of the twist $A_i$.

Unlike decoupled linear approximations~\cite{li2025hunyuan, chen2025deepverse, li2025vmem} that update translation and rotation independently, our formulation jointly integrates linear and angular velocities directly on the $SE(3)$ manifold. This design yields geometrically precise camera trajectories even under complex user actions involving tightly coupled translation and rotation.

\subsection{Camera-Controlled Video Generation}
\label{sec:camera_control}
We design a video generative model conditioned on the camera motion derived from the actions. Given the sequence of relative camera poses
$\{\Delta P_i\}_{i=1}^{F}$ which are obtained via action-to-camera mapping (Section~\ref{sec:action_control}), we accumulate them into global camera poses aligned with the first frame of the window. To provide explicit view-dependent geometric conditioning,
the camera poses are converted into Plücker embeddings
$\hat{P} \in \mathbb{R}^{F \times 6}$~\cite{sitzmann2021light}.

To inject camera control into the DiT,
we introduce a lightweight camera embedding module $c_\phi$ consisting of two MLP layers. Since the VAE temporally compresses the input by a factor of $r$,
we align camera conditioning with the latent sequence by concatenating
$r$ consecutive Plücker embeddings for each latent frame,
resulting in $\hat{\mathbf{p}} \in \mathbb{R}^{f \times (6r)}$, where $f = F / r$.
The camera embeddings are added to the DiT features $\mathbf{d}$ after each self-attention layer:
\begin{equation}
\mathbf{d} \leftarrow \mathbf{d} + c_\phi(\hat{\mathbf{p}}).
\label{eq:camera_emb}
\end{equation}

\subsection{Pose-Anchored Long-Term Memory}
\label{sec:3d_consistency}
Beyond precise action control, we leverage the camera pose derived in Section~\ref{sec:action_control} as a geometric anchor to maintain 3D consistency, ensuring spatial coherence when revisiting locations or viewpoints.

\vspace{-5pt} 
\paragraph{\normalfont\bfseries Global pose accumulation.}
Since our action-to-camera mapping strictly adheres to $SE(3)$ geometry, relative motions can be reliably accumulated into global camera poses. Specifically, the global pose of the $j$-th frame $I_j$ is computed via pose composition:
\begin{equation}
{P}_j^{\text{global}} = {P}_{j-1}^{\text{global}} \circ \Delta {P}_j, \quad {P}_0^{\text{global}} = \mathbf{I},
\end{equation}
where $\mathbf{I}$ is the identity and $\circ$ denotes pose composition.

\vspace{-5pt}
\paragraph{\normalfont\bfseries Pose-indexed memory retrieval.}
To exploit long-horizon spatial context, we maintain a long-term memory pool $\mathcal{M}$ that stores previously generated latents with their global camera poses. Due to the temporal compression of the video VAE, each latent $z_j$ corresponds to a short clip of $r$ consecutive frames and is therefore associated with a set of $r$ global camera poses. For simplicity, we denote each memory entry as $(P_j^{\text{global}}, z_j)$, where $P_j^{\text{global}}$ represents the set of global camera poses associated with the $r$ frames:
\begin{equation}
P_j^{\text{global}} =
\begin{bmatrix}
R_j & t_j \\
\mathbf{0}^\top & 1
\end{bmatrix}.
\end{equation}

Regarding the complex layouts and frequent occlusions in our gaming environments, we adopt a hierarchical memory retrieval strategy that uses the global camera pose as a spatial index over the memory bank. We first select the top-$K$ candidates $\mathcal{M}_{\text{trans}}$ whose camera positions are closest to the current position ${t}_i$:
\begin{equation}
\mathcal{M}_{\text{trans}} =
\operatorname{TopK}_{K}
\left(
- \lVert {t}_j - {t}_i \rVert_2
\;;\;
(P_j^{\text{global}}, z_j) \in \mathcal{M}
\right).
\end{equation}

From $\mathcal{M}_{\text{trans}}$, we further select $L$ entries ($L \le K$) whose viewing directions are most aligned with the current orientation ${R}_i$, measured using the trace of the relative rotation matrix:
\begin{equation}
\mathcal{M}_{\text{rot}} =
\operatorname{TopK}_{L}
\left(
\operatorname{tr}({R}_j^\top {R}_i)
\;;\;
(P_j^{\text{global}}, z_j) \in \mathcal{M}_{\text{trans}}
\right).
\end{equation}

\vspace{-5pt}
\paragraph{\normalfont\bfseries Long-term memory conditioning.}
The retrieved memory entries $(P_j^{\text{global}}, z_j) \in \mathcal{M}_{\text{rot}}$ provide the geometric context required to enforce spatial consistency during generation. We concatenate the retrieved latents with the current input latent sequence. Their associated camera poses are realigned to the first frame of the current denoising window, embedded via the camera embedding module $c_\phi$, and injected into the intermediate features of the DiT (Equation~\ref{eq:camera_emb}). This allows the model to establish explicit geometric correspondences between the current latents and long-term memory latents.

\begin{figure*}[!t]
    \centering
    \includegraphics[width=1\textwidth]{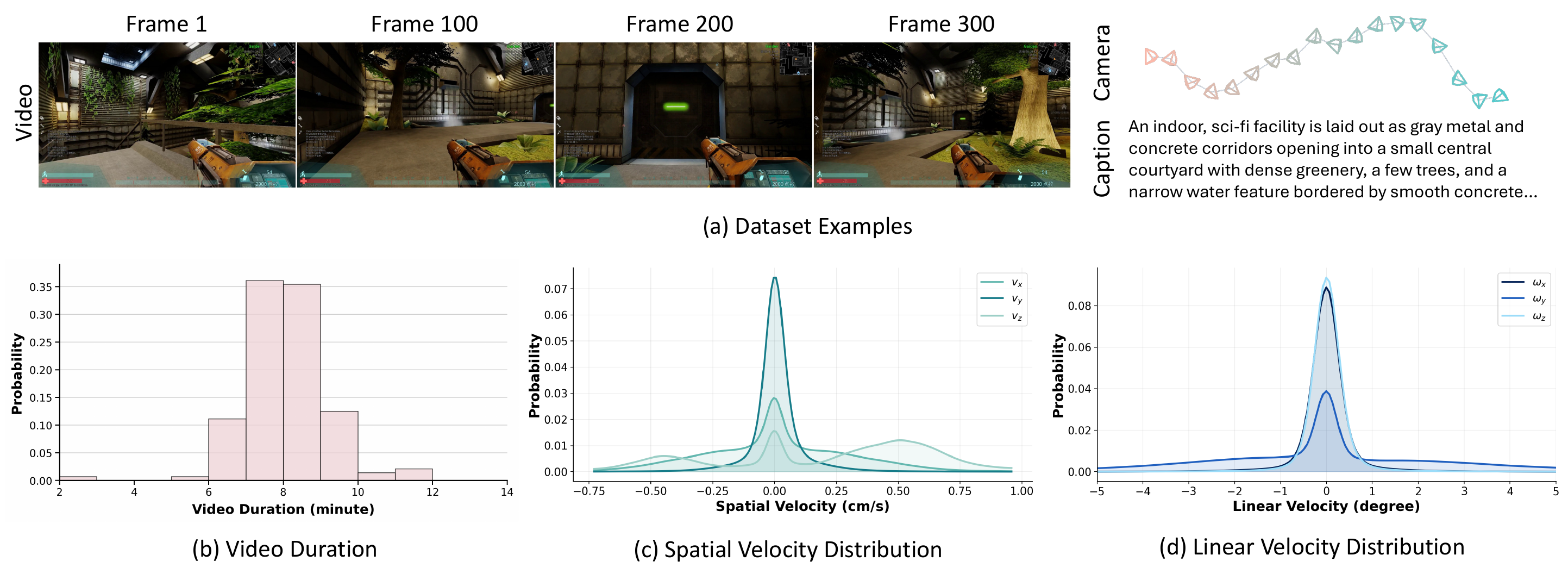}
    \vspace{-20pt}
    \caption{\textbf{Dataset samples and statistics:} 
    (a) Example gameplay frames annotated with camera trajectories, and text captions. 
    (b) Distribution of video durations. 
    (c) Distribution of linear velocities $(v_x, v_y, v_z)$. 
    (d) Distribution of angular velocities $(\omega_x, \omega_y, \omega_z)$. 
    The dataset captures diverse and authentic human gameplay behaviors for training interactive gaming world models.}
    \label{fig:dataset}
\end{figure*}

\subsection{Progressive Autoregressive Inference}
\label{sec:autoregressive}

\paragraph{\normalfont\bfseries Progressive noise scheduling.}
To support long-horizon autoregressive video diffusion, we adopt a progressive per-frame noise schedule that assigns monotonically increasing noise levels to latent frames within each denoising window. This design provides a reliable low-noise anchor in early frames, while keeping future frames at higher noise levels and thus correctable. This enables efficient cross-frame conditioning with large window overlap and stable rollout~\cite{xie2025progressive, chen2024diffusion, chen2025skyreels}. During training, conditioning on partially noisy context improves robustness to corrupted frames and reduces the train–test mismatch that would otherwise amplify error accumulation at inference time.

Specifically, we discretize the diffusion process into $N$ inference steps with monotonically increasing noise levels and partition them into $S$ stages, where $S$ equals the number of noisy latent frames maintained in the sequence.
The noisy latent sequence $\mathbf{z}_t$ (Section~\ref{sec:baseline}) is reformulated as a stage-wise $\mathbf{z}_s$, where $s \in \{0, \ldots, S-1\}$ denotes the noise stage, and each latent frame must complete all denoising steps across all stages to be fully denoised.
After completing all $S$ stages, the latent sequence is shifted forward: the earliest latent frame is evicted and decoded by the VAE, and a newly initialized pure-noise latent frame is appended to the end of the sequence.

\vspace{-5pt}
\paragraph{\normalfont\bfseries Attention sink.}
While progressive noise improves temporal stability, it can still accumulate errors under large and complex gaming motions, leading to visual saturation and distorted UI elements. To mitigate this, we incorporate an attention sink mechanism inspired by StreamingLLM~\cite{xiao2023efficient}, which stabilizes attention by anchoring a small set of initial tokens. During inference, we retain global initial frames as attention anchors, helping preserve frame fidelity, scene style, and UI consistency.

\vspace{-5pt}
\paragraph{\normalfont\bfseries Short-term memory.}
In practice, we find that providing recently generated latents, referred to as short-term memory, is crucial for reducing error drift during autoregressive generation.
We empirically set the number of short-term memory latents to match the number of generated latents, striking a balance between stability and computational efficiency.

\section{WorldCam-50h}
Large-scale interactive gaming datasets that capture authentic human action dynamics are essential for training foundational gaming world models, yet remain challenging to collect. Prior works~\cite{xiao2025worldmem, he2025matrix, guo2025mineworld} often rely on the Minecraft dataset~\cite{guss2019minerl}, which offers limited visual diversity, or on closed-licensed gameplay videos~\cite{li2025hunyuan}, which hinder reproducible research. 

To address these limitations, we introduce \textbf{WorldCam-50h}, a large-scale dataset of human gameplay videos. The dataset is annotated with detailed textual descriptions and camera pose information. Dataset samples and statistics are shown in Figure~\ref{fig:dataset}. Data preprocessing pipeline is described in Appendix~\ref{sec:preprocessing}.

\vspace{-5pt}
\paragraph{\normalfont\bfseries Data collection.}
We record human gameplay from one closed-licensed game, \textit{Counter-Strike}\footnote{\url{https://www.counter-strike.net}.}, and two open-licensed titles, \textit{Xonotic}\footnote{\url{https://xonotic.org}.} and \textit{Unvanquished}\footnote{\url{https://unvanquished.net}.}, chosen for their complex 3D environments and high interactivity. Note that all game-derived figures and video clips in the paper and supplementary material are from \textit{Xonotic} and \textit{Unvanquished}, licensed under CC BY-SA 2.5 and GPL v3. We focus on single-player exploration of static environments without dynamic objects or other players. To capture realistic human action dynamics, participants are instructed to perform diverse behaviors, including navigation, combined keyboard–mouse inputs, rapid camera movements, and revisiting locations (Figure~\ref{fig:dataset}(c) \& (d)). We collect over 100 videos per game, each averaging 8 minutes (Figure~\ref{fig:dataset}(b)), totaling about 1,000 minutes ($\approx$17 hours) of gameplay per game.

\vspace{-5pt}
\paragraph{\normalfont\bfseries Captioning.}
While prior works~\cite{bruce2024genie, po2025long, gao2025adaworld} often discard textual captions during training, we find textual guidance essential for maintaining frame quality and scene style. We therefore generate detailed captions for each training video chunk using Qwen2.5-VL-7B~\cite{yang2025qwen3}. Specifically, we prompt the model with: \textit{``Describe the static world of this video in one concise paragraph. Focus on the global layout (overall topology, primary regions, spatial arrangement, and key objects), the visual theme (colors, materials, and architectural style), and the ambient environmental conditions (overall lighting and weather).''}

\vspace{-5pt}
\paragraph{\normalfont\bfseries Camera annotation.}
We further extract global camera pose annotations using ViPE~\cite{huang2025vipe} to estimate both camera intrinsics and extrinsics for each one-minute segment of gameplay video. Since ViPE can produce erroneous estimates (e.g., unrealistically large translations), we apply additional filtering based on a predefined threshold on translation magnitudes.

\section{Experiments}
\label{sec:experiments}

\subsection{Implementation Details}
\label{sec:implementation}
We use Wan2.1-1.3B-T2V~\cite{wan2025wan} as our video DiT backbone. The spatial resolution is $480 \times 832$. During training, we use 8 progressive latents, 8 short-term memory latents, and 4 long-term memory latents. All experiments are conducted on 8 NVIDIA H100 GPUs.

Training is performed in three stages: 
(1) camera-controlled video generation with short-term memory for 10k iterations with a batch size of 64; 
(2) progressive autoregressive training with short-term memory for 10k iterations with a batch size of 48; and 
(3) progressive autoregressive training with both short- and long-term memory for 10k iterations with a batch size of 16 to enforce 3D consistency.

For progressive autoregressive training, we use $N = 64$ and $S = 8$, indicating that each latent is denoised for 8 sampling steps per stage, resulting in a total of 64 sampling timesteps across 8 stages. We use the AdamW optimizer with a learning rate of $1\times10^{-5}$ for all stages.

\subsection{Evaluation Settings} We compare our method with state-of-the-art interactive gaming world models, including Yume~\cite{mao2025yume}, Matrix-Game 2.0~\cite{he2025matrix}, and GameCraft~\cite{li2025hunyuan}, as well as camera-controlled video generation models, including CameraCtrl~\cite{he2024cameractrl} and MotionCtrl~\cite{wang2024motionctrl}. Additional evaluation details for the compared baselines are provided in \ref{sec:appendix_comparison}.

We evaluate three key aspects: action controllability, visual quality, and 3D consistency. To evaluate complex action controllability, we randomly sample 70 action trajectories across 70 starting images from the test split of our dataset. We generate 200-frame sequences for interactive gaming world models (Table~\ref{tab:quantitative_all}) and 16-frame sequences for camera-controlled methods, since they do not support long-horizon inference and suffer from severe visual drift under window-based autoregressive rollout (Table~\ref{tab:short_term_action}). For interactive gaming world models, action inputs are carefully mapped to each model's corresponding trained action space. For camera-controlled models, which require explicit camera trajectories, we provide ground-truth camera poses.

For visual quality and 3D consistency, we randomly select 50 starting images from our test dataset and predefine four simple closed-loop trajectories per image, resulting in 200 evaluation cases. We generate 200-frame sequences for interactive gaming world models (Table~\ref{tab:long_horizon_consistency}). Here, camera-controlled methods are not included, as they do not support long-horizon video generation and therefore cannot be evaluated for long-horizon 3D consistency.

\begin{figure*}[!t]
    \centering
    \includegraphics[width=1.0\textwidth]{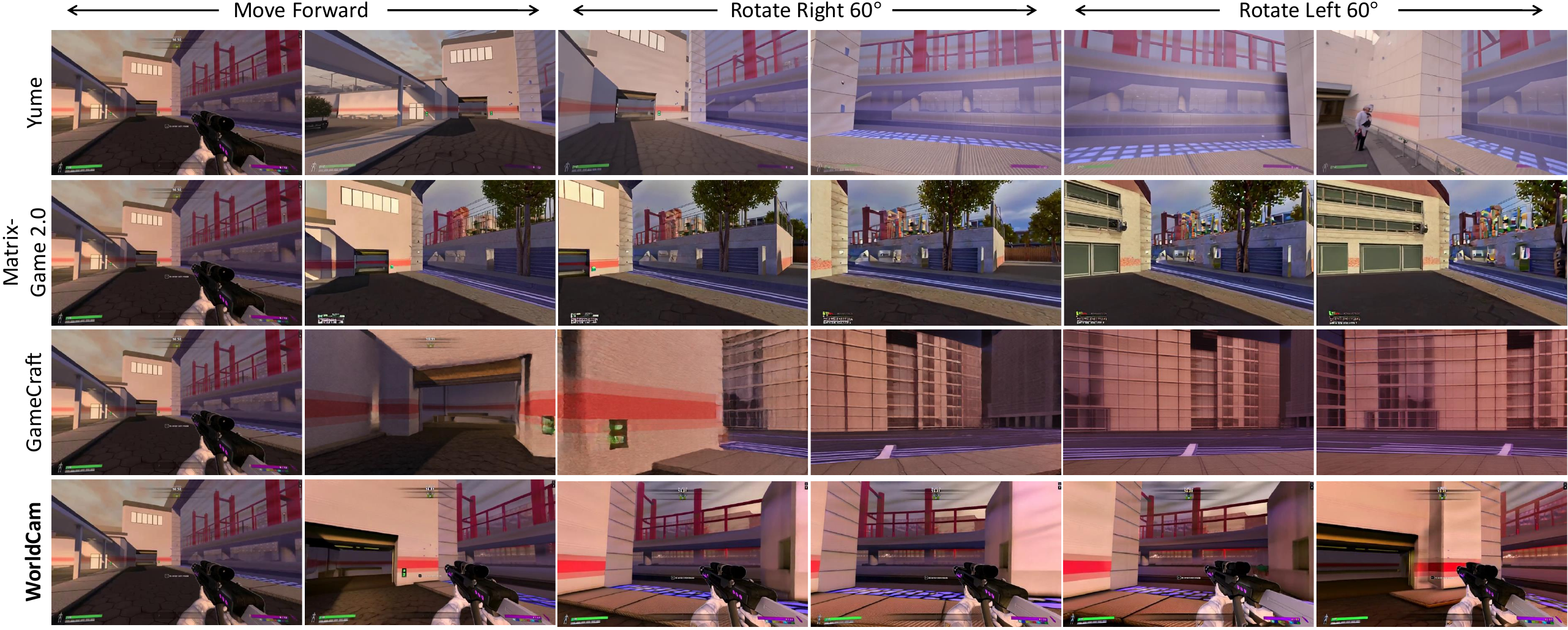}
    \vspace{-15pt}
    \caption{\textbf{Qualitative comparison with recent interactive gaming world models:} Compared to prior works, WorldCam faithfully follows user actions and maintains coherent 3D scene structure with high visual fidelity over long horizons.}
    \label{fig:main_qual}
\end{figure*}

\subsection{Evaluation Metrics}
\paragraph{\normalfont\bfseries Action controllability.}
We evaluate how accurately generated videos follow input actions using average Relative Pose Errors in translation (RPE$_{\text{trans}}$), rotation (RPE$_{\text{rot}}$), and camera extrinsics (RPE$_{\text{camera}}$) between camera trajectories estimated from generated videos by ViPE~\cite{huang2025vipe} and ground-truth camera motion. We apply Sim(3) Umeyama alignment~\cite{umeyama2002least} to compensate for differences in scale and coordinate frames.

\vspace{-5pt}
\paragraph{\normalfont\bfseries Visual quality.} We evaluate long-horizon visual quality using VBench++~\cite{huang2024vbench++} metrics. We report Aesthetic Quality (Aesth.), Subject Consistency (Subj. Cons.), Background Consistency (Bg. Cons.), Imaging Quality (Img.), Temporal Flickering (Temp.), and Motion Smoothness (Motion.), along with the average score (Avg.) across all categories.

\begin{table*}[t]
\centering
\footnotesize
\caption{
\textbf{Quantitative comparison with recent interactive gaming world models}
in terms of action controllability and visual quality.
All methods are evaluated under the same action trajectories with 200-frame video generation.
}
\label{tab:quantitative_all}
\vspace{-5pt}
\resizebox{\textwidth}{!}{
\begin{tabular}{l|ccc|ccccccc}
\toprule
 & \multicolumn{3}{c|}{Action Controllability}
 & \multicolumn{7}{c}{Visual Quality} \\
\cmidrule(lr){2-4} \cmidrule(lr){5-11}
Method
 & RPE$_{\text{trans}}\downarrow$
 & RPE$_{\text{rot}}(^\circ)\downarrow$
 & RPE$_{\text{camera}}\downarrow$
 & Avg.$\uparrow$
 & Aesth.$\uparrow$
 & Subj.\ Cons.$\uparrow$
 & Bg.\ Cons.$\uparrow$
 & Img.$\uparrow$
 & Temp.$\uparrow$
 & Motion.$\uparrow$ \\

\midrule
Yume~\cite{mao2025yume}
 & 0.111 & 2.222 & 0.137
 & 0.774 & \underline{0.476} & 0.741 & \underline{0.892} & 0.600 & 0.955 & \textbf{0.986} \\

Matrix-Game 2.0~\cite{he2025matrix}
 & 0.098 & 1.656 & 0.119
 & 0.766 & 0.457 & 0.741 & 0.843 & \underline{0.633} & 0.937 & 0.981 \\

GameCraft~\cite{li2025hunyuan}
 & \underline{0.086} & \underline{1.146} & \underline{0.100}
 & \underline{0.781} & 0.464 & \underline{0.804} & 0.850 & 0.626 & \underline{0.958} & \textbf{0.986} \\

\midrule
\textbf{WorldCam}
 & \textbf{0.080} & \textbf{0.696} & \textbf{0.086}
 & \textbf{0.844} & \textbf{0.508} & \textbf{0.896} & \textbf{0.959} & \textbf{0.752} & \textbf{0.964} & \underline{0.984} \\

\bottomrule
\end{tabular}}
\end{table*}

\begin{wraptable}[9]{r}{0.50\textwidth}
    \caption{\textbf{Quantitative results for long-horizon 3D consistency.}}
    \label{tab:long_horizon_consistency}
    \centering
    \resizebox{\linewidth}{!}{
    \begin{tabular}{l|cccc|c}
        \toprule
        Method
        & PSNR$\uparrow$
        & LPIPS$\downarrow$
        & MEt3R$\downarrow$
        & DINO Sim.$\uparrow$
        & Sharpness$\uparrow$ \\
        \midrule
        Real Videos  & - & - & - & - & 577 \\
        \midrule
        Yume~\cite{mao2025yume} & \underline{16.03} & {0.5629} & 0.0905 & 0.4545 & 95 \\
        Matrix-Game 2.0~\cite{he2025matrix} & 13.66 & \underline{0.4997} & {0.0662} & \underline{0.6153} & 179 \\
        GameCraft~\cite{li2025hunyuan} & {14.27} & 0.5749 & \underline{0.0489} & {0.5960}  & \underline{201} \\
        \midrule
        \textbf{WorldCam} & \textbf{16.69} & \textbf{0.3277} & \textbf{0.0342} &  \textbf{0.8884}& \textbf{656} \\
        \bottomrule
    \end{tabular}}%
\end{wraptable}
\vspace{-5pt}
\paragraph{\normalfont\bfseries 3D consistency.} To evaluate whether the generated environment maintains geometric consistency over long horizons, we measure view consistency using PSNR and LPIPS between all intermediate palindromic frame pairs in 200-frame closed-loop camera trajectories.

To account for the potential effect of blur on pixel-level metrics, we additionally report Sharpness, measured as the variance of the Laplacian~\cite{pertuz2013analysis}, averaged over all frames in each video. Blurry outputs (low sharpness) tend to yield artificially high PSNR and low LPIPS due to reduced pixel variance. 

We further report MEt3R~\cite{asim2025met3r}, which measures geometric multi-view consistency by reconstructing dense 3D point maps with DUSt3R~\cite{wang2024dust3r} between consecutive frames, warping DINO features~\cite{caron2021emerging} across views via the estimated geometry, and computing their similarity. 

In addition, we report DINO Similarity, defined as the cosine similarity between DINOv2~\cite{oquab2023dinov2} features of palindrome-corresponding frame pairs, directly measuring appearance consistency at the same camera pose.

\vspace{-5pt}
\paragraph{\normalfont\bfseries Human evaluation.}
Due to the absence of comprehensive benchmarks for interactive gaming world models, we supplemented our quantitative results with a user study focusing on action controllability, visual quality, and 3D consistency. We recruited 30 participants, each assessing 20 test cases drawn randomly from a large evaluation pool. The evaluation was conducted in a blind and randomized manner. Participants rated each model on a scale of 1 (worst) to 5 (best) based on detailed evaluation criteria.

\subsection{Comparison}

\begin{wraptable}[8]{r}{0.47\textwidth}
    \caption{\textbf{Comparison with camera-controlled models under 16-frame generation.}}
    \label{tab:short_term_action}
    \centering
    \resizebox{\linewidth}{!}{
    \begin{tabular}{l|ccc}
        \toprule
        Method
        & RPE$_{\text{trans}}\downarrow$
        & RPE$_{\text{rot}}(^\circ)\downarrow$
        & RPE$_{\text{camera}}\downarrow$ \\
        \midrule
         CameraCtrl~\cite{he2024cameractrl}             & \underline{0.071} & \underline{0.943} & \underline{0.083}\\
        MotionCtrl~\cite{wang2024motionctrl}  & 0.080& 1.271 & 0.102 \\
        \midrule
        \textbf{WorldCam}                        & \textbf{0.026} & \textbf{0.386} & \textbf{0.030} \\
        \bottomrule
    \end{tabular}}%
\end{wraptable}

\paragraph{\normalfont\bfseries Quantitative results.}
Table~\ref{tab:quantitative_all} and ~\ref{tab:long_horizon_consistency} summarize the quantitative comparison between our method and state-of-the-art interactive gaming world models~\cite{mao2025yume, he2025matrix, li2025hunyuan}. Our approach consistently outperforms existing baselines in terms of action controllability, visual quality, and 3D consistency on 200-frame long-horizon sequences by a large margin.

In Table~\ref{tab:quantitative_all}, while prior methods relying on text prompts~\cite{mao2025yume}, raw action signals~\cite{he2025matrix}, or linearly approximated cameras~\cite{li2025hunyuan} suffer from suboptimal action accuracy, our model achieves the lowest errors across all pose metrics, with RPE$_{\text{camera}}$ improved by 16.3\% (from 0.086 for the second-best GameCraft to 0.100). In Table~\ref{tab:short_term_action}, our method achieves the best camera accuracy in short-horizon comparisons with camera-controlled methods, improving RPE$_{\text{camera}}$ by 36.1\% (0.030 vs.\ 0.083 for the second-best CameraCtrl).

In terms of long-horizon visual quality, our method achieves the highest VBench average score (0.844), outperforming the second-best GameCraft (0.781) by a relative margin of 8.1\%. This improvement is attributed to progressive noise scheduling, attention sinks, and longer short-term latents, further analyzed in Section~\ref{sec:ablation_study}.

Regarding 3D consistency in Table~\ref{tab:long_horizon_consistency}, our method achieves the best performance on all evaluation metrics. While Yume~\cite{mao2025yume} achieves comparable PSNR, this is primarily due to blurrier outputs, as indicated by its substantially lower Sharpness score (95 vs.\ 577 for real videos). Overall, our method demonstrates the strongest 3D consistency across all metrics.

\begin{figure*}[!t]
    \centering
    \includegraphics[width=1\textwidth, height=0.90\textheight, keepaspectratio]{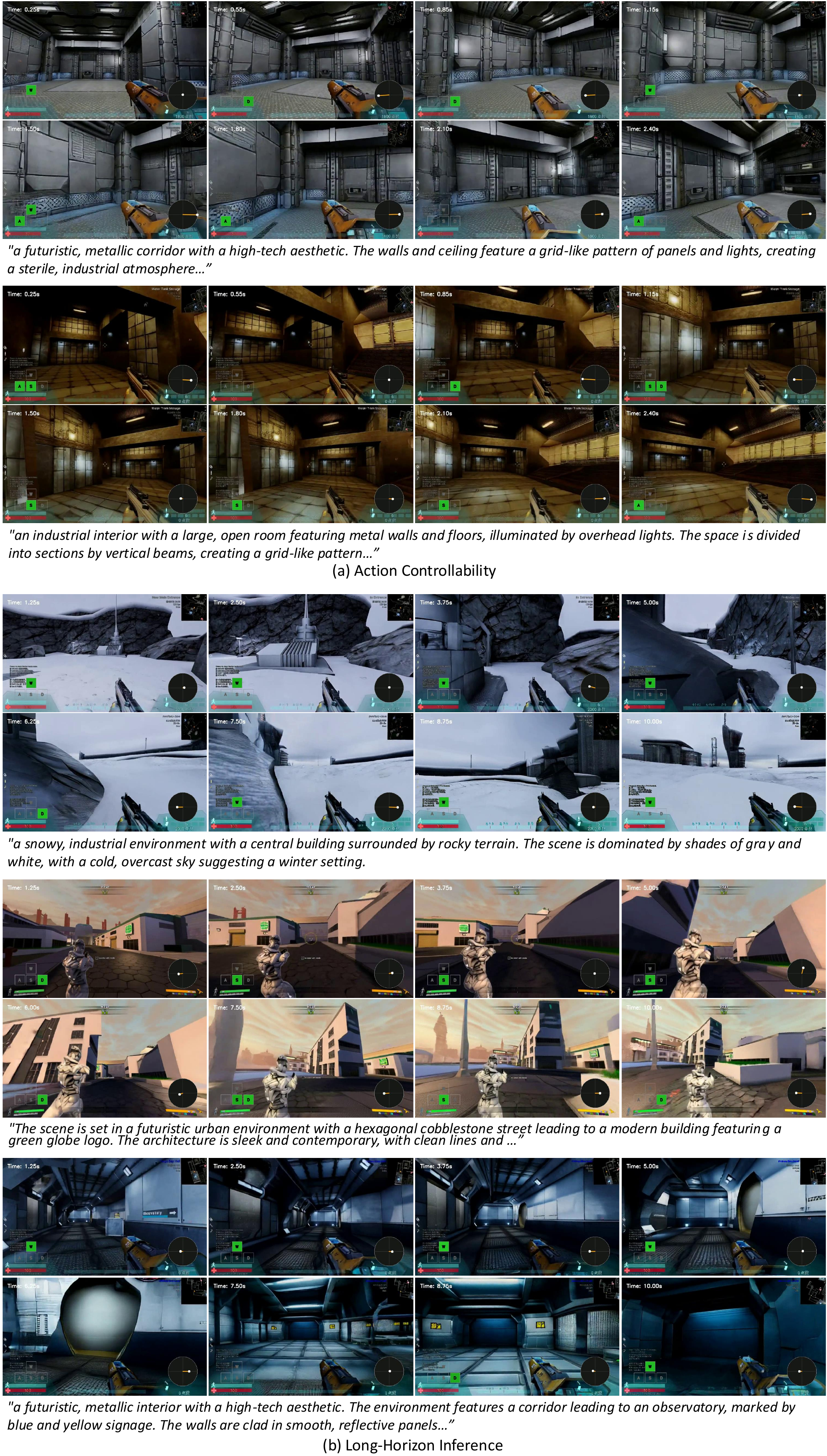}
    \caption{\textbf{Qualitative Results (Best viewed in color and zoomed in):} (a) WorldCam enables fine-grained action control coupled with simultaneous keyboard and mouse inputs. (b) WorldCam generates long-horizon videos exceeding 10 seconds (20 FPS) without error drift. Time (seconds at 20 FPS) is visualized in the top-left of each frame, while keyboard and mouse inputs are shown in the bottom-left and bottom-right, respectively.}
    \label{fig:supp_qual_1}
\end{figure*}

\begin{figure*}[!t]
    \centering
    \includegraphics[width=1\textwidth, height=0.90\textheight, keepaspectratio]{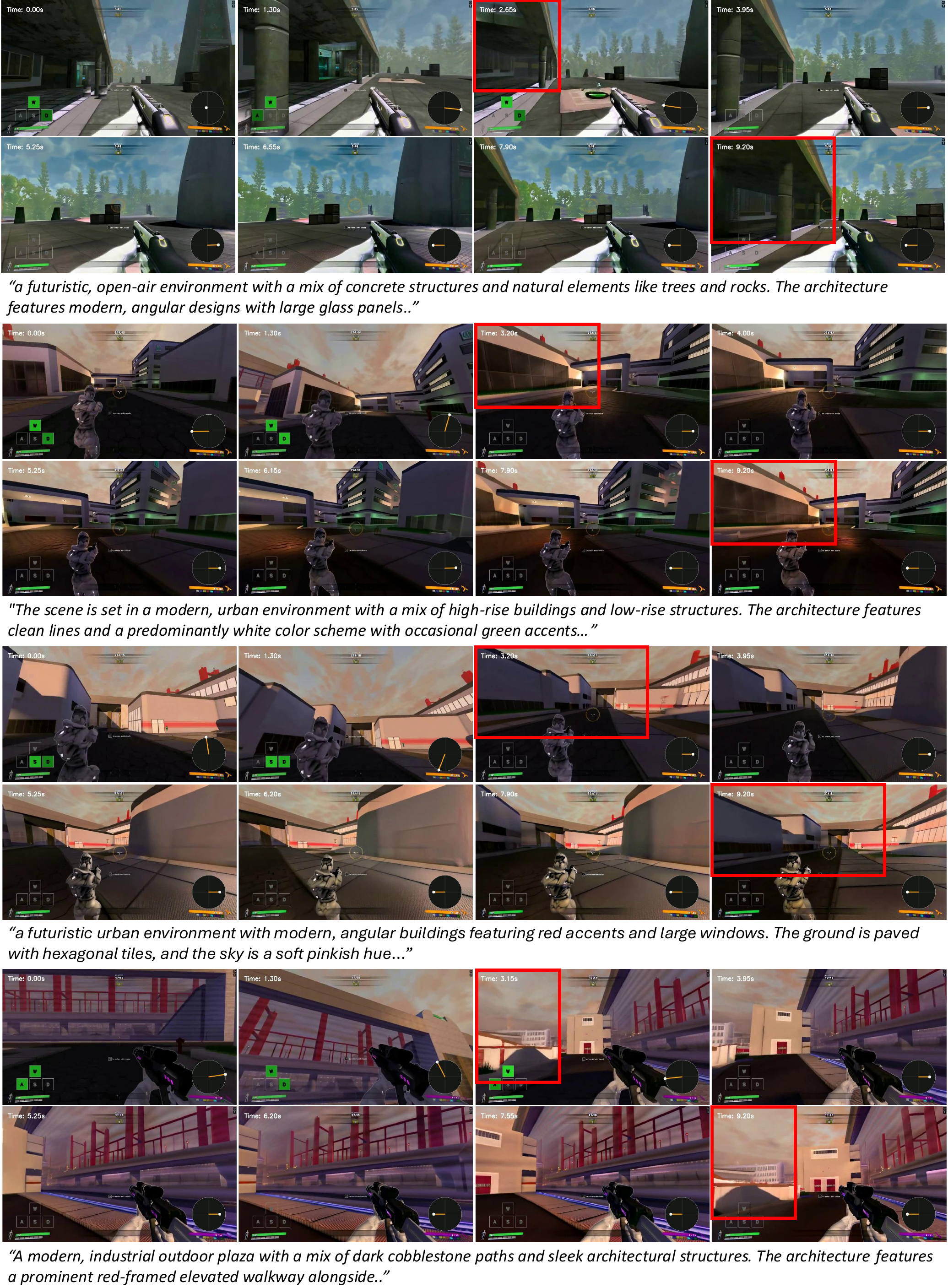}
    \caption{\textbf{Qualitative Results (Best viewed in color and zoomed in):} Our model preserves the underlying 3D scene consistently over long-horizon beyond the denoising window. Time (seconds at 20 FPS) is visualized in the top-left of each frame, while keyboard and mouse inputs are shown in the bottom-left and bottom-right, respectively. The red box highlights consistent 3D geometry in revisited views.}
    \label{fig:supp_qual_2}
\end{figure*}

\vspace{-5pt}
\paragraph{\normalfont\bfseries Qualitative results.}
Figures~\ref{fig:main_qual}, \ref{fig:supp_qual_1}, and \ref{fig:supp_qual_2} present qualitative results. As shown in Figure~\ref{fig:main_qual}, our method accurately follows user actions while maintaining high visual quality and consistent 3D geometry over long horizons. In contrast, prior methods often struggle with challenging action inputs, exhibit quality degradation over time, and fail to preserve 3D consistency beyond the denoising window. 

Figure~\ref{fig:supp_qual_1}(a) demonstrates that our method effectively handles challenging coupled keyboard and mouse inputs, while Figure~\ref{fig:supp_qual_1}(b) shows that it can generate plausible long-horizon outputs with newly emerging scenes while remaining faithful to user actions. Figure~\ref{fig:supp_qual_2} further highlight that our method preserves consistent 3D geometry when users revisit previously seen locations and viewpoints beyond the denoising window.

\begin{wraptable}[6]{r}{0.42\textwidth}
\vspace{-15pt}
    \caption{\textbf{Human evaluation results.} We report average user ratings
on a scale of 1 (worst) to 5 (best).}
\vspace{-5pt}
    \label{tab:user_study}
    \centering
    \resizebox{0.9\linewidth}{!}{
        \begin{tabular}{l|ccc}
            \toprule
            Method & \shortstack{Action\\Controllability}$\uparrow$ & \shortstack{Visual\\Quality}$\uparrow$ & \shortstack{3D\\Consistency}$\uparrow$ \\
            \midrule
            Yume~\cite{mao2025yume}            & 2.47 & 2.83 & 1.44 \\
            Matrix-2.0~\cite{he2025matrix} & \underline{3.78} & \underline{3.42} & 2.75 \\
            GameCraft~\cite{li2025hunyuan}       & 2.55 & 3.34 & \underline{3.36} \\
            \midrule
            \textbf{WorldCam}           & \textbf{4.31} & \textbf{4.44} & \textbf{4.36} \\
            \bottomrule
        \end{tabular}}%
\end{wraptable}

\vspace{-5pt}
\paragraph{\normalfont\bfseries Human evaluation.}
Table~\ref{tab:user_study} further validates the effectiveness of our method across all three evaluation aspects, where our approach consistently outperforms prior methods by a clear margin, achieving relative improvements of 14.0\%, 29.8\%, and 29.8\% over the second-best baselines in action controllability, visual quality, and 3D consistency, respectively.

\subsection{Ablation Study}
\label{sec:ablation_study}
\begin{table}[h]
\centering
\footnotesize
\begin{minipage}[t]{0.54\textwidth}
\centering
\caption{\textbf{Ablation on the number of long-term memory latents for visual quality and 3D consistency.}}
\vspace{-5pt}
\label{tab:long_memory_3d_abl}
\resizebox{\linewidth}{!}{
\begin{tabular}{c|ccccccc|cc}
\toprule
\#
& Avg.$\uparrow$
& Aesth.$\uparrow$
& Subj.\ Cons.$\uparrow$
& Bg.\ Cons.$\uparrow$
& Img.$\uparrow$
& Temp.$\uparrow$
& Motion.$\uparrow$
& PSNR$\uparrow$
& LPIPS$\downarrow$ \\
\midrule
0 &   0.840 & \textbf{0.511} & {0.876} & {0.955}& {0.751} &\textbf{0.961} &\textbf{0.984} & 12.163& 0.591 \\
1 &  {0.840}& \textbf{0.511} &0.879 & 0.955& \textbf{0.752}& \textbf{0.961}&\textbf{0.984}& 12.624 &0.573 \\
4 & \textbf{0.841} & 0.510 &  \textbf{0.881} & \textbf{0.956} & 0.750& \textbf{0.961}& \textbf{0.984} & \textbf{12.950}	&\textbf{0.554} \\
\bottomrule
\end{tabular}}%
\end{minipage}%
\hfill
\begin{minipage}[t]{0.44\textwidth}
\centering
\caption{\textbf{Ablation on the number of short-term memory latents for visual quality.}}
\vspace{-6pt}
\label{tab:memory_abl}
\resizebox{\linewidth}{!}{
\begin{tabular}{c|ccccccc}
\toprule
\#
& Avg.$\uparrow$
& Aesth.$\uparrow$
& Subj.\ Cons.$\uparrow$
& Bg.\ Cons.$\uparrow$
& Img.$\uparrow$
& Temp.$\uparrow$
& Motion.$\uparrow$
\\
\midrule
1 &  0.749&  0.317& 0.873 & 0.947 &0.414 & 0.959& 0.983  \\
4 & {0.836} & \textbf{0.514} &  0.873&  0.947&  0.737 &0.959&0.983\\
8 &  \textbf{0.840} & {0.511} & \textbf{0.876} & \textbf{0.955}& \textbf{0.751} &\textbf{0.961} &\textbf{0.984} \\
\bottomrule
\end{tabular}}%
\end{minipage}
\end{table}

\paragraph{\normalfont\bfseries Component analysis.} 
Table~\ref{tab:long_memory_3d_abl} demonstrates that increasing the number of long-term memory latents improves PSNR and SSIM, thereby enhancing 3D consistency. Under our progressive noise scheduling, different noise levels require distinct historical contexts, consequently, a larger long-term memory provides richer cues to preserve spatial structures. Notably, visual quality remains stable even in the absence of long-term memory ($0.841$ Avg. with 4 frames vs. $0.840$ without), indicating the model's robustness in generating new scenes without severe error drift, even when spatial constraints are relaxed. 

\begin{wraptable}[6]{r}{0.58\textwidth}
\caption{\textbf{Ablation on the effect of attention sinks for visual quality.}}
\label{tab:attn_sink_abl}
\centering
\resizebox{\linewidth}{!}{
\begin{tabular}{c|ccccccc}
\toprule
\Method
& Avg.$\uparrow$
& Aesth.$\uparrow$
& Subj.\ Cons.$\uparrow$
& Bg.\ Cons.$\uparrow$
& Img.$\uparrow$
& Temp.$\uparrow$
& Motion.$\uparrow$
\\
\midrule
w/o attention sink &  {0.840} & \textbf{0.511} & {0.876} & {0.955}& {0.751} &{0.961} &\textbf{0.984}  \\
with attention sink &   \textbf{0.841}& 0.502 & \textbf{0.883} &  \textbf{0.956}&\textbf{0.753} & \textbf{0.964} &  \textbf{0.984}  \\
\bottomrule
\end{tabular}}%
\end{wraptable}

Tables~\ref{tab:memory_abl} and~\ref{tab:attn_sink_abl} ablate the number of short-term latents and the inclusion of an attention sink regarding long-horizon visual quality. As discussed in Section~\ref{sec:autoregressive}, increasing short-term latents and introducing an attention sink reduce long-horizon error drift, ultimately improving the Avg. score.

\paragraph{\normalfont\bfseries Action-to-camera mapping.}
\begin{wraptable}[9]{r}{0.48\textwidth}
\caption{\textbf{Ablation study on action-to-camera mapping.}}
\label{tab:camera_control}
\centering
\resizebox{\linewidth}{!}{
\begin{tabular}{l|ccc}
\toprule
Method 
& RPE$_{\text{trans}}\downarrow$
& RPE$_{\text{rot}}(^\circ)\downarrow$
& RPE$_{\text{camera}}\downarrow$ \\
\midrule
Yume~\cite{mao2025yume} & 0.111 & 2.222 & 0.137 \\
Matrix-Game 2.0~\cite{he2025matrix} & 0.098 & 1.656 & 0.119 \\
GameCraft~\cite{li2025hunyuan} & \underline{0.086} & {1.146} & \underline{0.100} \\
\midrule
WorldCam (Linear) & {0.093} &	\underline{0.962} &	{0.102} \\
\textbf{WorldCam (Lie)} & \textbf{0.080} & \textbf{0.696} & \textbf{0.086} \\
\bottomrule
\end{tabular}}
\end{wraptable}
Table~\ref{tab:camera_control} compares linear and Lie algebra-based approximations for action-to-camera mapping in our model. WorldCam (Linear) achieves performance comparable to GameCraft~\cite{li2025hunyuan}, while improving the rotational error. WorldCam (Lie) further improves all metrics 
and achieves the best overall performance. These results demonstrate that our method provides strong action controllability, and that the Lie algebra-based approximation more accurately models camera motion than the linear approximation.

\paragraph{\normalfont\bfseries Memory retrieval.}
\begin{wraptable}{r}{0.35\textwidth}
    \caption{\textbf{Ablation study on long-term memory retrieval.}}
    \label{tab:memory_retrieval}
    \centering
    \resizebox{0.8\linewidth}{!}{
    \begin{tabular}{l|ccc}
        \toprule
        Method
        & PSNR$\uparrow$
        & LPIPS$\downarrow$
        & MEt3R$\downarrow$ \\
        \midrule

        Random & \underline{15.76} & \underline{0.3645} & 0.041 \\
        Temporal & 15.18 & 0.3867 & \underline{0.040} \\

        \midrule
        \textbf{Ours} & \textbf{16.42} & \textbf{0.3496} & \textbf{0.038} \\
        \bottomrule
    \end{tabular}}%
\end{wraptable}

Table~\ref{tab:memory_retrieval} compares our memory retrieval strategy with random and temporal retrieval. The former randomly selects long-term latents from the memory pool, while the latter retrieves the most recent latents. Our camera pose-based retrieval significantly outperforms both baselines.  For this ablation study, we evaluate on 100 generated videos, each consisting of 200 palindrome frames.

Note that our framework is agnostic to the retrieval strategy. In this work, we adopt a lightweight method based on camera position and orientation. Field-of-View (FOV) overlap is a natural alternative used in prior works~\cite{yu2025context}, which typically assumes constant depth to unproject a generated 2D image into 3D and reproject it to a target camera pose to estimate viewpoint overlap. However, this approximation becomes unreliable in complex game environments with frequent occlusions and narrow corridors. Accurate FOV computation would require precise depth maps for generated frames, introducing additional overhead. Therefore, to keep retrieval efficient and compatible with our latent space memory design, we adopt a camera pose-based retrieval strategy. 

\section{Conclusion}
We introduce WorldCam, a foundational interactive gaming world model that enables precise action control, long-horizon generation, and consistent 3D world modeling by using camera pose as a unifying geometric anchor for both immediate action control and long-horizon 3D world consistency. We also present WorldCam-50h, a large-scale dataset of human gameplay that includes open-licensed games, supporting research reproducibility. Extensive experiments demonstrate clear improvements over prior interactive world models and camera-controlled baselines in terms of action accuracy, long-horizon visual quality, and 3D consistency.

\clearpage

\appendix
\section*{Appendix Overview}

\begin{itemize}
    \renewcommand\labelitemi{$\bullet$}
    \item Section~\ref{sec:preprocessing}: Data preprocessing pipeline.
    \item Section~\ref{sec:appendix_comparison}: Detailed comparison with existing works and evaluation details for each comparison baseline.
    \item Section~\ref{sec:abl_action_to_camera}: Detailed explanation of linear and Lie algebra-based action-to-camera approximation.
    \item Section~\ref{sec:limitation}: Limitations and runtime comparison.
\end{itemize}\

\section{Data Preprocessing}
\label{sec:preprocessing}
We segment the collected videos (with an average duration of 8 minutes) into 1-minute clips (1,800 frames at 30 FPS) and extract global camera extrinsic and intrinsic parameters for each clip using ViPE~\cite{huang2025vipe}. The 1,800-frame clips provide sufficient temporal context for selecting long-term memory frames. Each 1,800-frame clip is further divided into segments of 64 frames, consisting of 32 frames for short-term memory and 32 frames for generation.

To train with long-term memory latents, long-term memory frames are selected from the same 1,800-frame clip based on camera translation and rotation similarity, following Section~\ref{sec:3d_consistency}. Specifically, using the camera pose of the last frame of the 64-frame segment as a spatial index, we select 4 long-term memory clips (4 frames each) from the remaining 1,736 frames, excluding the current 64-frame segment. To reduce training time, we precompute all VAE latents and store them in a cache.

\section{Comparison Details}
\label{sec:appendix_comparison}

\subsection{Comparison with Interactive Gaming World Models}
\label{sec:game_world_models}

In this section, we highlight the novelty of our approach and provide evaluation details along with comparisons to recent interactive gaming world models, including Matrix-Game 2.0~\cite{he2025matrix}, GameCraft~\cite{li2025hunyuan}, and Yume~\cite{mao2025yume}. As summarized in Table~\ref{tab:model} of the main paper, our model satisfies three key requirements for an interactive 3D gaming world model: precise action control, 3D consistency, and long-horizon inference.

Many prior works focus on embedding raw action inputs directly into the model (The Matrix~\cite{feng2024matrix}, Matrix-Game~\cite{zhang2025matrix}, Matrix-Game 2.0~\cite{he2025matrix}, Genie~\cite{bruce2024genie}, Yan~\cite{ye2025yan}). However, these approaches cannot ensure 3D consistency, since raw actions cannot be directly used as spatial indices for memory retrieval. GameCraft~\cite{li2025hunyuan} employs camera poses for action control, but it does not leverage them to maintain long-term 3D consistency. Camera-controlled video generation models (CameraCtrl~\cite{he2024cameractrl}, MotionCtrl~\cite{wang2024motionctrl}) can achieve short-horizon 3D consistency (\eg, within 16 frames), but they are not designed for world modeling, which limits both action control and long-horizon inference.

Consequently, none of the prior works satisfy all three requirements for interactive gaming world models. In contrast, we adopt camera poses as a unified representation for both precise action control and long-horizon 3D consistency. 

\subsection{Evaluation Details of Comparison Works}
\noindent\textbf{Matrix-Game 2.0}~\cite{he2025matrix} is trained on an internal 1,200-hour dataset collected from Unreal Engine and GTA5 environments. Keyboard actions ($W, A, S, D$) are embedded via cross-attention mechanisms, while mouse movements ($dx, dy$) are injected into intermediate feature representations. The model adopts a few-step distillation strategy based on a causal architecture to enable low-latency streaming video generation.

For evaluation, we convert our action space $(v_x, v_z, \omega_x, \omega_y)$ into the Matrix-Game 2.0 pre-trained action space $(W, A, S, D, dx, dy)$. Since Matrix-Game does not control the speed of each action, we carefully select thresholds to activate each action. Specifically, linear velocities $v_x$ and $v_z$ are thresholded to remove noise and mapped to binary key inputs $W, S$ (forward/backward) and $A, D$ (left/right), while angular velocities $\omega_x$ and $\omega_y$ are thresholded and converted into mouse movements $dy$ and $dx$, respectively.

While Matrix-Game 2.0 directly injects raw action inputs into the generative model, our method maps input actions to camera poses and leverages these poses for precise action control and long-horizon 3D consistency.

\noindent\textbf{GameCraft}~\cite{li2025hunyuan} employs a chunk-wise autoregressive video diffusion model with a convolutional camera embedder that encodes Pl\"ucker embeddings and injects them into video latents. The model is trained on more than 100 AAA games and synthetic datasets, with camera poses extracted using Monst3r~\cite{zhang2024monst3r}. During inference, discrete actions (\eg, forward, backward, left, right, turn left/right, tilt up/down) and their associated speeds are converted into camera poses via linear approximations.

For evaluation, we map our action space $(v_x, v_z, \omega_x, \omega_y)$ into their predefined action space (\eg, mapping $\omega_y$ to turn left/right) and convert the corresponding velocities into speeds via scaling. We carefully select a speed scaling factor to match our velocity magnitudes with the speed settings used in the pre-trained GameCraft model.

While GameCraft performs chunk-wise generation, our approach adopts autoregressive progressive noise scheduling for latent-by-latent generation, enabling interactive world modeling. Moreover, GameCraft relies on linear approximations for action-to-camera mappings, whereas our method introduces a Lie algebra-based action-to-camera mapping for fine-grained action control and further leverages predicted camera poses to support long-horizon world consistency.

\noindent\textbf{Yume}~\cite{mao2025yume} employs a chunk-wise autoregressive masked video diffusion transformer trained on the Sekai-Real-HQ dataset~\cite{li2025sekai}. Camera motion is discretized into predefined actions (\eg, forward, backward, left, right, turn left/right, still) and incorporated into text captions. To maintain spatial consistency, Yume adopts FramePack~\cite{zhang2025packing}, which compresses historical latents via downsampling.

For evaluation, since the model generates chunks of eight latents at a time conditioned on text that encodes actions, we average our actions across the eight latents, apply thresholding, and convert them into Yume's predefined textual action space (\eg, mapping $v_z$ to ``\textit{Person moves forward}''). Because Yume encodes actions through text prompts, the resulting action control is relatively coarse. We attempt to match the motion speed by specifying speed-related cues in the text prompts.

While Yume encodes actions through textual prompts and compresses historical latents using FramePack, our model is trained directly on camera poses derived from action inputs and retrieves only relevant historical latents from a long-term memory pool.

\section{Action-to-Camera Approximation}
\label{sec:abl_action_to_camera}

In this section, we provide a detailed comparison between linear and Lie algebra-based action-to-camera approximations. The key difference lies in whether translation and rotation are updated independently or jointly integrated on the \textit{SE}(3) manifold. This distinction is particularly important for interactive 3D world modeling, where user actions often induce coupled camera motions.

Let the action at time step \(i\) be represented as a 6-DoF velocity vector
\begin{equation}
A_i = 
\begin{bmatrix}
\mathbf{v}_i \\
\boldsymbol{\omega}_i
\end{bmatrix}
=
\begin{bmatrix}
v_x, v_y, v_z, \omega_x, \omega_y, \omega_z
\end{bmatrix}^{\top},
\end{equation}
where \(\mathbf{v}_i \in \mathbb{R}^3\) and \(\boldsymbol{\omega}_i \in \mathbb{R}^3\) denote the linear and angular velocities, respectively.

\paragraph{\normalfont\bfseries Linear approximation.} 
A common approximation used in prior works~\cite{li2025hunyuan, li2025hunyuan} is to update translation and rotation separately. Given the camera pose
\(
P_{i-1} =
\begin{bmatrix}
R_{i-1} & t_{i-1} \\
0^\top & 1
\end{bmatrix}
\in SE(3),
\)
the linear approximation computes
\begin{equation}
t_i = t_{i-1} + R_{i-1}\mathbf{v}_i,
\qquad
R_i = R_{i-1}\exp(\widehat{\boldsymbol{\omega}}_i),
\label{eq:linear_update}
\end{equation}
where \(\widehat{\boldsymbol{\omega}}_i \in \mathfrak{so}(3)\) is the skew-symmetric matrix of \(\boldsymbol{\omega}_i\). The resulting pose is then
\begin{equation}
P_i =
\begin{bmatrix}
R_i & t_i \\
0^\top & 1
\end{bmatrix}.
\end{equation}

Although simple, this formulation updates translation and rotation independently, failing to capture coupled rigid-body motions. Under screw motion (\eg moving forward while turning right), the camera should follow a curved trajectory induced by simultaneous translation and rotation, whereas the linear approximation applies these updates separately, leading to accumulated trajectory drift.

\paragraph{\normalfont\bfseries{Lie algebra-based approximation.}}
In contrast, our formulation represents the action directly as a twist in \(\mathfrak{se}(3)\). Specifically, we construct
\begin{equation}
\widehat{A}_i =
\begin{bmatrix}
\widehat{\boldsymbol{\omega}}_i & \mathbf{v}_i \\
0^\top & 0
\end{bmatrix}
\in \mathfrak{se}(3),
\end{equation}
and compute the relative camera motion via the matrix exponential
\begin{equation}
\Delta P_i = \exp(\widehat{A}_i)
=
\begin{bmatrix}
\Delta R_i & \Delta t_i \\
0^\top & 1
\end{bmatrix}.
\label{eq:lie_exp}
\end{equation}
The camera pose is then updated as
\begin{equation}
P_i = P_{i-1}\Delta P_i.
\label{eq:lie_update}
\end{equation}

Unlike Equation~\ref{eq:linear_update}, Equation~\ref{eq:lie_update} jointly integrates linear and angular velocities on the \(SE(3)\) manifold. Consequently, the translational component \(\Delta t_i\) is not treated as a simple vector addition, but as part of a geometrically consistent rigid transformation coupled with rotation. This allows the model to accurately represent screw motions and other complex camera trajectories arising from simultaneous translation and rotation.

\begin{figure}[h]
\centering

\begin{minipage}{0.52\textwidth}
\centering
\includegraphics[width=\linewidth]{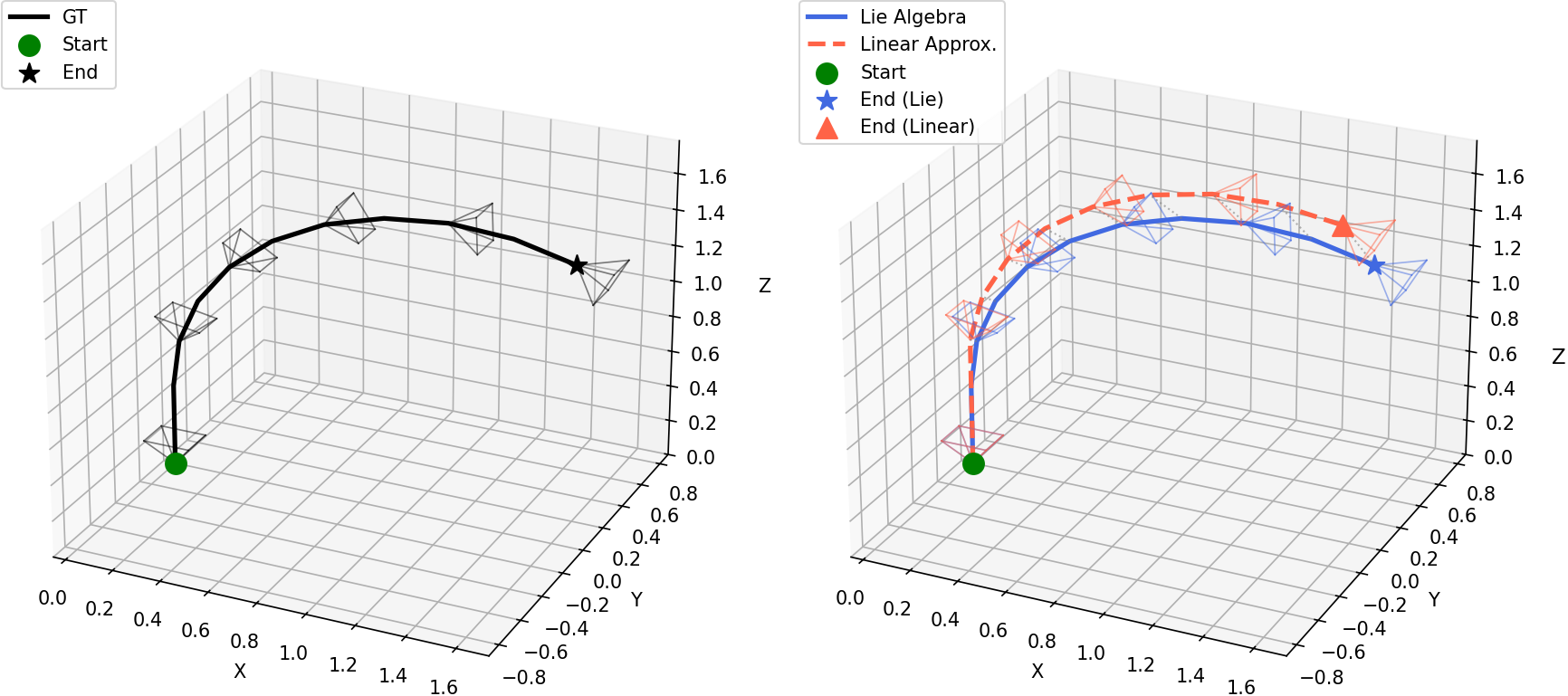}
\caption{Visualization of the comparison between linear and Lie algebra-based action-to-camera mappings.}
\label{fig:action_camera}
\end{minipage}
\hfill
\begin{minipage}{0.44\textwidth}
\centering
\captionof{table}{\textbf{Comparison between linear and Lie algebra-based action-to-camera approximations.} 
All values are scaled by $10^{3}$ for better visibility.}
\label{tab:theoretical_comparison}

\resizebox{0.9\linewidth}{!}{
\begin{tabular}{l|ccc}
\toprule
Method
& $\text{RPE}_{\text{trans}} \downarrow$
& $\text{ATE}_{\text{avg}} \downarrow$
& $\text{ATE}_{\text{final}} \downarrow$ \\
\midrule
Linear & 0.564 & 29.239 & 91.149 \\
\textbf{Lie algebra} & \textbf{0.001} & \textbf{0.005} & \textbf{0.015} \\
\bottomrule
\end{tabular}}
\end{minipage}

\end{figure}
\paragraph{\normalfont\bfseries{Why screw motion matters.}}
This difference is visualized in Figure~\ref{fig:action_camera}, which considers a simple screw motion example: moving forward while turning right. In this case, the ground-truth camera trajectory follows a smooth curved path. The linear approximation fails to reproduce this motion because it decouples translation from rotation, resulting in inaccurate relative motion and accumulated trajectory drift. In contrast, the Lie algebra-based formulation closely matches the ground-truth trajectory, since it preserves the coupling between translation and rotation at every step.

Table~\ref{tab:theoretical_comparison} quantitatively confirms this behavior. For evaluation, we generate 50 trajectories of length 200 frames and compute Relative Pose Error (\(\text{RPE}_{\text{trans}}\)) and Absolute Trajectory Error (\(\text{ATE}\)). Specifically, \(\text{ATE}_{\text{avg}}\) and \(\text{ATE}_{\text{final}}\) measure the drift between the approximated and ground-truth camera trajectories over the full sequence and at the final frame, respectively. Linear approximations exhibit large \(\text{RPE}\) and accumulated \(\text{ATE}\), indicating inaccurate relative camera motion and substantial long-term drift. In contrast, the Lie algebra--based approximation maintains near-zero error across all metrics. These results show that our formulation provides a stable and accurate camera representation, which is crucial for precise action control as well as stable long-term memory retrieval and spatial consistency (Section~\ref{sec:3d_consistency}).

\section{Limitation}
\label{sec:limitation}
While our approach improves generation quality and controllability, inference efficiency remains a limitation. This work primarily focuses on camera pose as a unifying geometric representation that jointly enables immediate action control and long-horizon, 3D consistent memory. Improving runtime is therefore orthogonal to our main contribution and could likely be achieved by integrating existing acceleration and distillation techniques. In particular, diffusion distillation methods~\cite{yin2025slow, geng2025mean, song2023consistency, yin2023one} could reduce the number of sampling timesteps per latent, potentially collapsing multi-step sampling into few-step generation, consistent with recent progress toward real-time video synthesis.

\begin{wraptable}[8]{r}{0.6\textwidth}
\caption{\textbf{Inference time comparison.} All methods are evaluated at a spatial resolution of \(480\times832\) on a single H100 GPU.}
\label{tab:runtime}
\centering
\resizebox{\linewidth}{!}{
\begin{tabular}{l|c|c|c|c|c}
\toprule
Method & Latent-wise AR & Steps & Frames/Chunk & Time/Chunk (s) & Time/Step (s) \\
\midrule
Yume~\cite{mao2025yume} & $\times$ & 4 & 32 & \textbf{2.91} & \underline{0.73} \\
GameCraft~\cite{li2025hunyuan} & $\times$ & 8 & 33 & 109.4 & 13.68 \\
\midrule
Matrix-Game 2.0~\cite{he2025matrix} & $\checkmark$ & 3 & 12 & \underline{3.09} & 1.03 \\
\textbf{WorldCam} & $\checkmark$ & 8 & 4 & 4.17 & \textbf{0.52} \\
\bottomrule
\end{tabular}}
\end{wraptable}

In Table~\ref{tab:runtime}, we compare the inference time complexity of our model with several recent approaches. For a fair comparison, all methods are evaluated at a spatial resolution of \(480\times832\) on a single H100 GPU. Yume~\cite{mao2025yume} and GameCraft~\cite{li2025hunyuan} are chunk-wise autoregressive models that generate 32 and 33 frames per chunk, respectively. While this design improves throughput per chunk, it is not well-suited for interactive world models that require low-latency interaction. In particular, chunk-wise generation requires a sequence of actions corresponding to all frames within a chunk before generation, which limits responsiveness in interactive settings.

In contrast, Matrix-Game 2.0~\cite{he2025matrix} and our method adopt latent-wise autoregressive generation, enabling low-latency interaction. Matrix-Game 2.0 generates three latent tokens every three sampling steps, whereas our method generates a single latent token using eight sampling steps. Notably, Matrix-Game 2.0 employs distillation to reduce the number of sampling steps, whereas our method does not use any distillation technique. Despite this, our method achieves a comparable time per chunk to Matrix-Game 2.0 while achieving the lowest time per step among all compared approaches. The low time per step suggests that further acceleration through distillation could potentially enable real-time interactive generation.

\clearpage

\clearpage
\newpage
\bibliographystyle{assets/plainnat}
\bibliography{ref}

\end{document}